\documentclass[sigconf]{acmart}

\usepackage{subcaption}
\usepackage{algorithm}
\usepackage{algorithmic}
\usepackage{tabularx}
\usepackage{lipsum} 
\usepackage{multirow}
\usepackage{xcolor}
\usepackage{tcolorbox}

\renewcommand\footnotetextcopyrightpermission[1]{} 

\acmConference[MMAsia '25]{ACM Multimedia Asia}{December 9--12, 2025}{Kuala Lumpur, Malaysia}
\acmBooktitle{Proceedings of the ACM Multimedia Asia (MMAsia '25), December 9--12, 2025, Kuala Lumpur, Malaysia}

\acmYear{2025}
\copyrightyear{2025}

\AtBeginDocument{%
  \renewcommand\footnotetextcopyrightpermission[1]{}
  \fancyfoot{}
  \fancyfoot[C]{\small This is the author-accepted version of a paper to appear in \textit{ACM Multimedia Asia 2025 (MMAsia '25)}. 
  The definitive version will be available at the ACM Digital Library.}
}

\begin{document}

\title{CharCom: Composable Identity Control for Multi-Character Story Illustration}

\author{Zhongsheng Wang}
\email{zhongsheng.wang@auckland.ac.nz}
\orcid{0009-0003-4235-7710}
\affiliation{%
  \institution{Wuhan University of Communication}
  \country{China}
}
\affiliation{%
  \institution{University of Auckland}
  \country{New Zealand}
}

\author{Ming Lin}
\email{mlin569@aucklanduni.ac.nz}
\orcid{0009-0006-6135-2861}
\affiliation{%
  \institution{Univerisity of Auckland}
  \country{New Zealand}
}

\author{Zhedong Lin}
\email{zlin629@aucklanduni.ac.nz}
\orcid{0009-0003-5079-9850}
\affiliation{%
  \institution{Univerisity of Auckland}
  \country{New Zealand}
}

\author{Yaser Shakib}
\email{Yaser@bedaia.ai}
\orcid{0009-0009-2783-7648}
\affiliation{%
  \institution{Bedaia.ai}
  \country{New Zealand}
}

\author{Qian Liu}
\email{liu.qian@auckland.ac.nz}
\orcid{0000-0002-3162-935X}
\affiliation{%
  \institution{Univerisity of Auckland}
  \country{New Zealand}
}

\author{Jiamou Liu}
\email{jiamou.liu@auckland.ac.nz}
\orcid{0000-0002-0824-0899}
\affiliation{%
  \institution{Univerisity of Auckland}
  \country{New Zealand}
}

\renewcommand{\shortauthors}{Zhongsheng Wang. et al.}

\begin{abstract}
Ensuring character identity consistency across varying prompts remains a fundamental limitation in diffusion-based text-to-image generation. We propose \textbf{CharCom}, a modular and parameter-efficient framework that achieves character-consistent story illustration through composable LoRA adapters, enabling efficient per-character customization without retraining the base model. Built on a frozen diffusion backbone, CharCom dynamically composes adapters at inference using prompt-aware control. Experiments on multi-scene narratives demonstrate that CharCom significantly enhances character fidelity, semantic alignment, and temporal coherence. It remains robust in crowded scenes and enables scalable multi-character generation with minimal overhead, making it well-suited for real-world applications such as story illustration and animation.
\end{abstract}


\begin{teaserfigure}
  \centering
  \includegraphics[width=\textwidth]{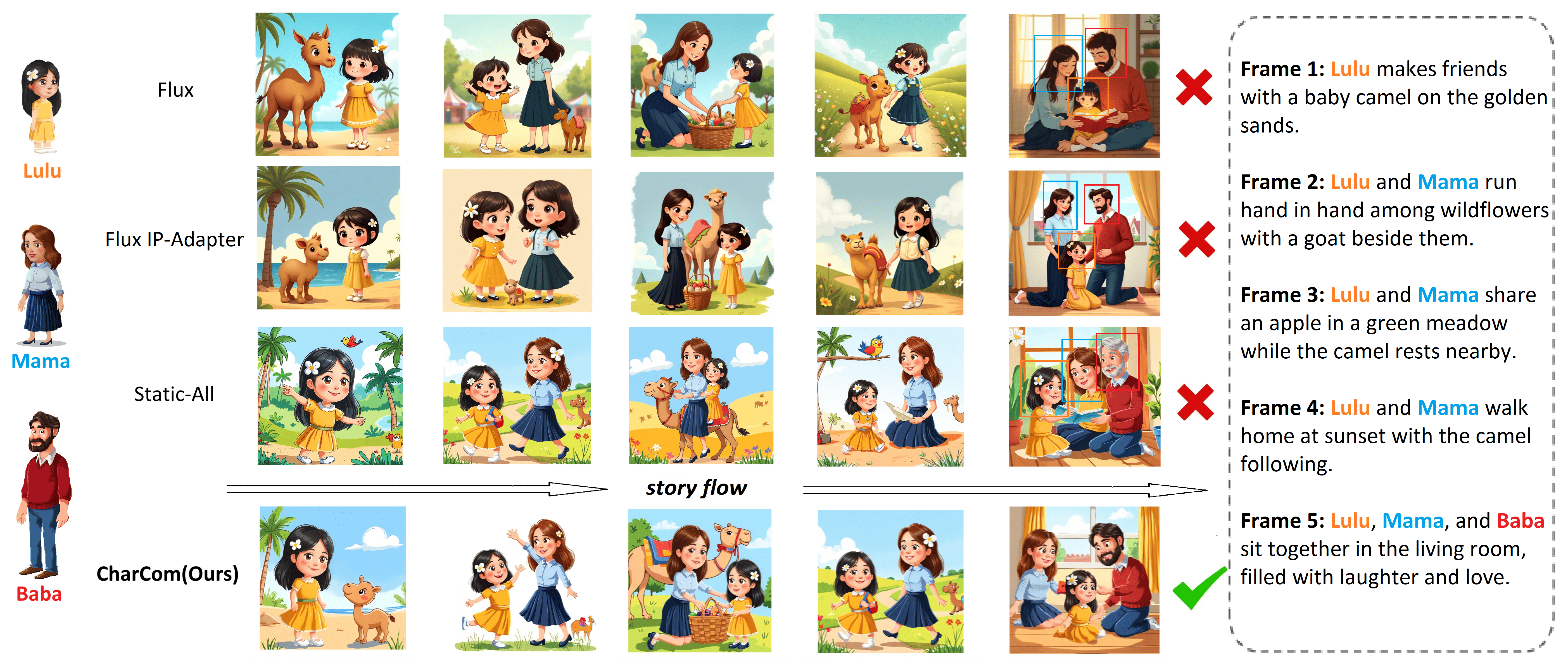}
  \caption{A set of illustration sequences generated using different numbers of character references, given a continuous story illustration description on the right. CharCom shows the best character consistency preservation comparing to other methods. The different colors in the fifth column of pictures frame the inconsistencies of different characters.}
  \Description{Enjoying the baseball game from the third-base
  seats. Ichiro Suzuki preparing to bat.}
  \label{fig:teaser}
\end{teaserfigure}


\maketitle
\section{Introduction}

Text-to-image (T2I) diffusion models have shown remarkable progress in generating photorealistic and semantically rich images from textual prompts. Mainstream diffusion models~\cite{rombach2022highresolution, gu2023assessing} are now widely used in creative domains such as visual storytelling, educational illustration, and children's books. However, despite their ability to generate high-resolution and visually vivid images, these models still struggle to maintain consistent character identity across multiple generated scenes~\cite{avrahami2023roleconsistency, zhou2024storymaker}.

This consistency is particularly critical for storybook illustration, where readers expect not only high visual quality but also temporal and semantic continuity of characters throughout a story. Current pipelines often suffer from \textit{identity drift}, where the same character appears with different facial features, hairstyles, clothing, or poses across scenes. Such inconsistencies disrupt narrative coherence and undermine the practical utility of diffusion models in real-world storytelling applications~\cite{maharana2022storydalleap, pan2022flintstonessv}. To mitigate character inconsistency, early approaches~\cite {ruiz2023dreambooth} perform full model fine-tuning for each subject to encode identity features, achieving high fidelity but incurring heavy computational overhead and poor scalability. More recently, IP-Adapter~\cite{ye2023ipadapter} introduces a plug-and-play visual adapter that projects identity features into the model's latent space. While efficient and training-free, it remains limited in complex narrative scenarios, particularly those involving multiple interacting characters or dynamic scene transitions~\cite{wang2024instantid, oracle2024oracle}. These limitations highlight the central challenge of our work: \textbf{\emph{Ensuring consistent visual identity for each character across a sequence of story scenes}}.

In this work, we propose \textbf{CharCom}, a lightweight, scalable, and composable framework for character-consistent story illustration using diffusion models. Inspired by Low-Rank Adaptation (LoRA)~\cite{hu2021lora}, our method represents each character with an independently trained rank-4 identity adapter. These adapters can be composed at inference time without modifying or retraining the backbone diffusion model, enabling multi-character scene generation with minimal training data and no fine-tuning overhead. To further enhance narrative coherence, we design a structured prompt template that explicitly controls key elements such as character roles, emotions, actions, and background context~\cite{cheng2024autostudiocc, tao2024coinal}. As shown in Figure~\ref{fig:teaser}, CharCom achieves more consistent character rendering and scene coherence than several popular methods. Visually, facial features remain highly stable, and characters retain their distinctive appearance across multiple scenes, offering a more faithful and engaging visual narrative.

We further introduce a character-aware evaluation framework based on GPT-4o~\cite{openai2024gpt4o}, which performs multimodal visual question answering (VQA) to assess identity fidelity and temporal coherence. Compared to conventional CLIP-based metrics~\cite{radford2021clip}, this approach offers more semantically grounded and human-aligned assessments~\cite{huang2023vqascore}. All experiments are conducted on \textit{Shakoo Maku}\footnote{https://www.shakoomaku.com/}, an Arabic-language animated IP for children aged 3--6 featuring recurring characters across scenes. To support evaluation, we construct a small-scale synthetic benchmark comprising 20 storylines, each with 5 prompts containing overlapping characters, generated with GPT-4o~\cite{hurst2024gpt} to ensure narrative diversity. CharCom demonstrates clear advantages over strong existing methods, achieving consistent generation with minimal computational overhead. Overall, our main contributions can be summarized as follows:
\begin{itemize}
    \item We propose \textbf{CharCom}, a lightweight framework that achieves consistent multi-character generation by stacking independently trained LoRA adapters without modifying the backbone diffusion model.
    \item We design a structured prompting scheme that encodes narrative details, such as roles, emotions, actions, and scenes, to enhance cross-frame coherence and controllability in storytelling tasks.
    \item We introduce a synthetic benchmark for character-consistent story illustration, enabling quantitative and qualitative comparison across models.
    \item Experiments show that CharCom outperforms strong existing methods across multiple narrative scenarios, providing a flexible solution for practical application environments.
\end{itemize}

\section{Related Work}

\noindent\textbf{Diffusion Models for T2I}

Text-to-image (T2I) diffusion models have advanced rapidly following the introduction of Denoising Diffusion Probabilistic Models (DDPMs)~\cite{ho2020ddpm}, which generate images by iteratively denoising Gaussian noise. DDIM~\cite{song2020ddim} further improves inference speed via a non-Markovian sampling process.

Latent diffusion models~\cite{rombach2022highresolution,gu2023ass} further improve efficiency by leveraging variational autoencoders (VAEs) to operate in latent space, significantly reducing memory and compute cost while enabling high-resolution generation. These models commonly adopt CLIP~\cite{radford2021clip} embeddings and classifier-free guidance to enhance prompt fidelity. FLUX~\cite{kang2025fluxak} introduces dual encoders and flow-matching loss for better prompt alignment under constrained compute budgets. In this work, we adopt FLUX as our base model and focus on enabling character consistency on top of this efficient backbone.

\subsection{Character Identity Preservation}

Maintaining consistent character identity across diverse prompts is a core challenge for diffusion-based models. Earlier works like DreamBooth~\cite{ruiz2023dreambooth} achieve high fidelity by fine-tuning the entire diffusion backbone with subject-specific prompts, but suffer from high computational cost and poor scalability. Textual Inversion~\cite{wei2025personalizedig} offers a more efficient alternative by optimizing subject-specific token embeddings while freezing the model backbone, though it often compromises visual fidelity and controllability.

To reduce fine-tuning overhead, Low-Rank Adaptation (LoRA)~\cite{hu2021lora,kumari2022multiconceptco} introduces a parameter-efficient strategy by decomposing weight updates into low-rank matrices. This method has been extended to support multi-concept blending~\cite{he2025conceptrolcc,wu2025lesstomoregu}, though these approaches typically require offline merging or scene-level planning to avoid interference among identities. Parallel to these, adapter-based inference methods such as IP-Adapter~\cite{ye2023ipadapter} and InstantID~\cite{wang2024instantid} offer training-free alternatives by injecting visual features through attention or facial embeddings. However, their conditioning mechanisms are often underconstrained in multi-character scenarios, leading to identity entanglement and reduced consistency.

\subsection{Multi-Character and Story-Level Modeling}

Recent efforts have extended T2I generation toward story-level consistency and multi-character coherence. OneActor~\cite{oneactor2023oneactor} uses pseudo-labels and compositional training to improve generalization, but is limited to single-subject scenarios. ORACLE~\cite{oracle2024oracle} enhances identity fidelity within LoRA-based generation through mutual information constraints, though this increases training complexity and limits scalability. More comprehensive systems such as StoryMaker~\cite{zhou2024storymaker} and CharacterFactory~\cite{wang2025characterfactory} aim to model entire narratives by integrating modules like scene decomposition, character disentanglement, or GAN-based identity embeddings.

However, these approaches rely on tightly coupled architectures or curated identity references, making them difficult to adapt across tasks or integrate into modular pipelines. Despite their progress, achieving both flexibility and consistency in multi-character, multi-scene generation remains an open challenge.

\section{Problem Definition}

The goal is \textit{to generate high-quality, semantically faithful, and temporally coherent multi-scene illustrations that preserve character identities across the entire narrative.}  
We consider a character set \( \mathcal{C} = \{c_1, \dots, c_n\} \), where each \( c_i \) is associated with a reference image set \( \mathcal{I}_{c_i} = \{ I_1^{(c_i)}, \dots, I_k^{(c_i)} \} \), and a prompt sequence \( \mathcal{P} = \{p_1, \dots, p_m\} \) describing multi-scene narratives. The task is to learn a generation function 
\(
\mathcal{G}_\Theta : \mathcal{P} \times \mathcal{C} \rightarrow \mathbb{R}^{H \times W \times 3}
\) that given a prompt \( p_j \) and a character subset \( \mathcal{S} \subseteq \mathcal{C} \), produces an image \( \hat{I}_{j}^{(S)} = \mathcal{G}_\Theta(p_j, S) \) of spatial resolution \( Height \times Weight \). The generated image should (1) preserve the identities of all characters in \( \mathcal{S} \) with respect to their references, (2) align semantically with the prompt \( p_j \), and (3) maintain temporal consistency across adjacent prompts in \( \mathcal{P} \).

The overall optimization objective is:
\begin{equation*}
\begin{aligned}
\mathcal{L}_{\text{total}} = \mathbb{E}_{(p,S)}\Big[ 
&\mathcal{L}_{\text{id}}\left( \mathcal{G}_\Theta(p, S), \{\mathcal{I}_c\}_{c \in S} \right) \\
&+ \lambda \cdot \mathcal{L}_{\text{sem}}\left( \mathcal{G}_\Theta(p, S), p \right) + \mu \cdot \mathcal{L}_{\text{temp}} 
\Big],
\end{aligned}
\end{equation*}
where \( \mathcal{L}_{\text{id}} \) measures identity preservation against the reference set, \( \mathcal{L}_{\text{sem}} \) encourages semantic alignment between the prompt and the generated image, and 
\(
\mathcal{L}_{\text{temp}} = \sum_{j=2}^m \mathcal{D}_{\text{id}}\left(\hat{I}_{j}^{(S)}, \hat{I}_{j-1}^{(S)}\right)
\) promotes consistency between temporally adjacent scenes.


\section{Methodology}

\begin{figure}[htbp]
  \centering
  \includegraphics[width=1\linewidth]{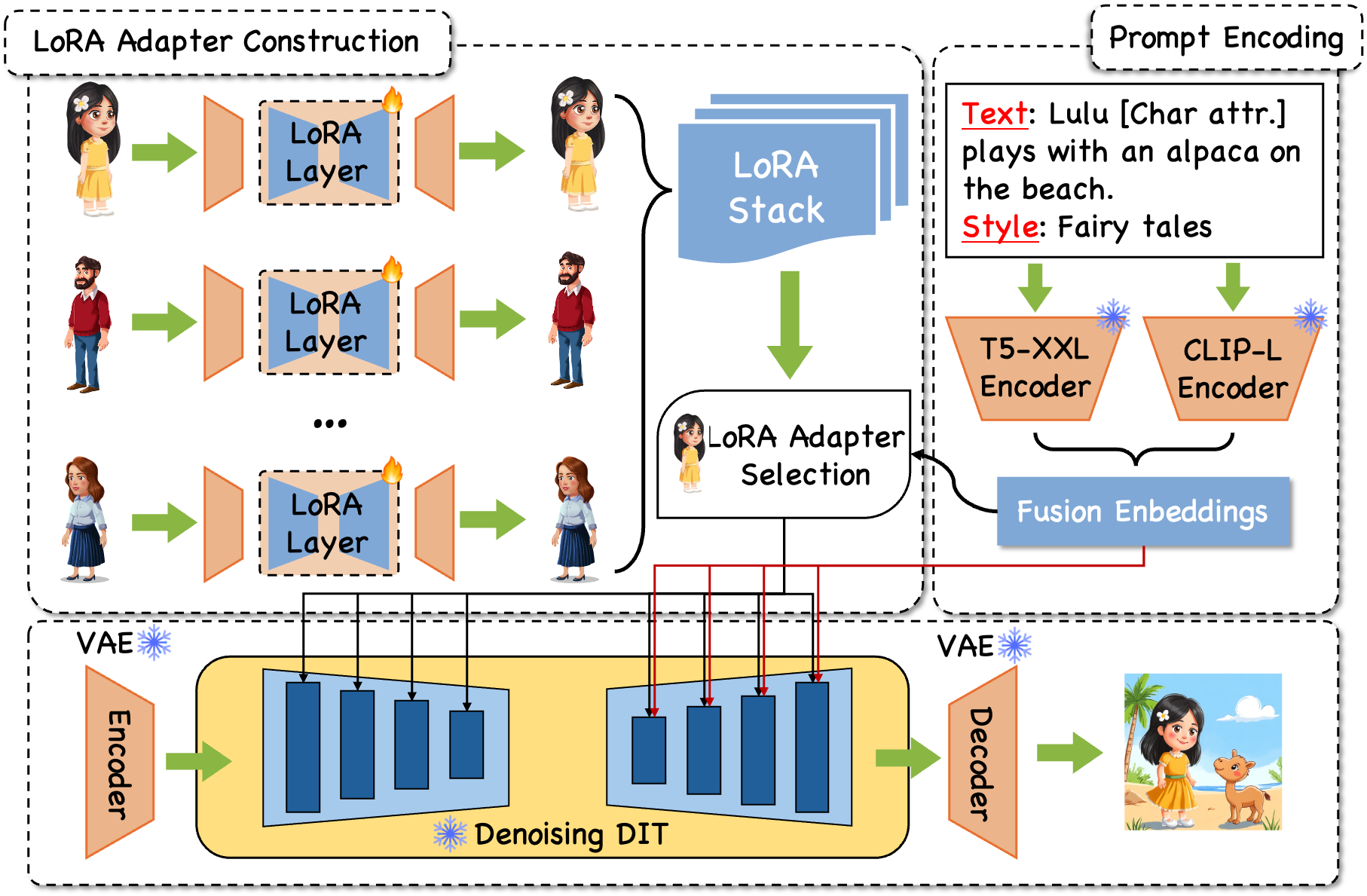}
  \caption{Overview of CharCom architecture built on FLUX~\cite{kang2025fluxak}. CLIP-L and T5-XXL encode prompts via dual encoders, and a frozen VAE handles image latents. Character-specific LoRA adapters are trained independently and composed at inference.}
  \label{fig:arch}
\end{figure}

We propose \textbf{CharCom}, a modular, composable framework for identity-consistent multi-character, multi-scene story illustration, built on top of FLUX~\cite{kang2025fluxak}, a recent diffusion-based text-to-image model. Unlike static conditioning methods such as IP-Adapter~\cite{ye2023ipadapter} or mutual-information-based LoRA systems like ORACLE~\cite{oracle2024oracle}, CharCom requires neither joint optimization nor backbone modifications. It trains lightweight, per-character adapters independently and composes them dynamically at inference, enabling flexible multi-character control without retraining. Formally, the model learns $\mathcal{G}_\Theta(p, S)$ to generate images that are temporally and semantically consistent given scene prompts and character identities. As illustrated in Figure~\ref{fig:arch}, CharCom runs atop the FLUX diffusion backbone, with per-character identity adapters selectively injected into attention and multi-layer perceptron layers at inference to enable controllable composition.

\subsection{System Pipeline}

The CharCom framework is composed of 4 main stages:
\begin{itemize}
\item \textbf{Few-shot Adaptation}: Given a reference image set $\mathcal{I}_{c}$ (e.g., 15–30 images) for each character $c$, we train a dedicated rank-4 LoRA adapter $\Delta\theta_c = \mathbf{B}_c \mathbf{A}_c$ while keeping the FLUX backbone frozen. Each adapter encodes character-specific residuals in the FLUX parameter space.

\item \textbf{Adapter Composition}: During image generation, for each prompt $p_j$ involving a character subset $\mathcal{S} \subseteq \mathcal{C}$, we construct $\Theta = \theta + \sum_{c \in \mathcal{S}} w_c \cdot \Delta\theta_c$ and compose the weighted adapters at inference time, enabling modular identity control within the diffusion model.

\item \textbf{Prompt Encoding}: We employ a dual-encoder setup with CLIP-L~\cite{radford2021learning} and T5-XXL~\cite{https://doi.org/10.48550/arxiv.2210.11416} to jointly encode prompts $\mathcal{P}$ into complementary visual and semantic representations, enhancing grounding and alignment.

\item \textbf{Latent Denoising}: A DiT-style U-Net denoiser is guided by FlowMatch~\cite{lipman2022flow} and a temporal consistency loss, enforcing coherent transitions across scenes.
\end{itemize}
The overall design enables efficient per-character training and scalable multi-character inference within a single-GPU memory budget.

\subsection{Low-Rank Adapter Training and Fusion}

We adopt low-rank adaptation~\cite{hu2021lora} to efficiently encode character-specific identity features into the diffusion model parameters. For any weight matrix \( W \in \mathbb{R}^{d \times d} \) in FLUX backbone, we insert a learnable low-rank residual:
\[
\Delta W_{c} = \mathbf{B_{c}}\mathbf{A_{c}}, \quad \text{where } \mathbf{A_{c}} \in \mathbb{R}^{d \times r},\ \mathbf{B_{c}} \in \mathbb{R}^{r \times d},\ r \ll d.
\]

During training, only \( \mathbf{A_{c}}, \mathbf{B_{c}} \) are updated, while all other backbone parameters remain frozen. For each character \( c \),  we optimize a reconstruction loss that measures the distance between the generated image and its corresponding reference target $y$, conditioned on a given prompt $x$:
\[
 \mathbb{E}_{(x, y)} \left\| f_{\theta + \mathbf{B}_c \mathbf{A}_c}(x) - y \right\|^2,
\]
where \( \theta \) denotes the frozen base parameters, and \( f_{\theta+\Delta W} \) represents the diffusion network evaluated at a noise scale \( \sigma_t \sim \mathcal{U}[0,1] \). Each LoRA adapter contains approximately 21k trainable parameters and can be trained within 1.5 hours on a single RTX 4090 GPU.

At inference, multiple character-specific adapters are combined through a weighted fusion mechanism:
\[
W^{\star} = W + \sum_{c \in \mathcal{S}} w_c \, \mathbf{B}_c \mathbf{A}_c,
\]
Here, \( w_c \in [0,1] \) reflects the prompt relevance for character $c$, computed via a dual-encoder similarity function:
\[
w_c = \sigma\left( \alpha \cdot \cos\left( f_{\text{T5}}(p_j), f_{\text{T5}}(\Phi_c) \right) + \beta \cdot \cos\left( f_{\text{CLIP}}(p_j), f_{\text{CLIP}}(\mathcal{I}_c) \right) \right),
\]
The fusion coefficients $\alpha$ and $\beta$ are set to 1.0 by default. For each prompt $p_j$, the similarity-based mechanism computes $w_c$ by aligning the prompt with the character’s textual description $\Phi_c$ and reference image set $\mathcal{I}_c$. This guides CharCom to automatically select and load only relevant LoRA adapters, while suppressing unrelated ones to avoid identity interference. The merging process incurs negligible overhead (<0.1 ms) and operates within a 20 GB VRAM budget.

\subsection{Consistency-Oriented Prompt Design}

To promote consistent visual outputs, we design structured prompts based on a hierarchical template
\[
\mathcal{P}_c = \underbrace{\tau_c}_{\text{trigger}} + \underbrace{\Gamma(\Phi_c)}_{\text{attributes}} + \underbrace{\mathcal{A}}_{\text{action}} + \underbrace{\mathcal{S}}_{\text{style}},
\]
where each component encodes distinct semantic factors to minimize cross-character interference. Here, \( \tau_c \) denotes a unique character-specific token (e.g., \texttt{Shakoo Maku Lulu}) used to explicitly identify each individual character. Each character is associated with an independently trained adapter, which is linked to its corresponding trigger token. Embedding these tokens directly into the prompt provides explicit identity grounding and enables the model to selectively activate the corresponding identity modules during inference. \( \Phi_c \) encodes fixed, character-specific attributes such as age, hairstyle, and clothing. We utilize Qwen2.5-Max~\cite{qwen25} to generate high-quality textual descriptions for each character that appears in \textit{Shakoo Maku} in advance. These descriptions are compressed into 15–25 token sequences via \( \Gamma(\cdot) \), balancing semantic richness with token efficiency. This compact representation ensures efficient processing within the attention mechanism. \( \mathcal{A} \) encodes task-specific actions (e.g., ``sits beside'', ``reads a book'') extracted from the input sentence, guiding the character's pose and interactions. In contrast, \( \mathcal{S} \) captures the global rendering style, including aesthetic cues such as the color palette and illustration type.

As shown in the example on the right, in multi-role scenarios, the prompts are concatenated in a canonical order. This structured prompt design encapsulates each character’s identity, attributes, and actions into localized semantic units, thereby reducing cross-character interference and promoting consistent identity preservation as well as coherent spatial layouts in multi-character scenes.

\section{Experiments}

\subsection{Experiment Setup}

\begin{tcolorbox}[float=!ht, colback=gray!5!white, colframe=gray, title=Compositional Prompt Example] \textbf{Characters(trigger+attributes):} \begin{itemize} \item Shakoo Maku Lulu: A little girl, likely around 5 or 6 years old, with a bright and gentle presence $\cdots$ \item Shakoo Maku Mama: A woman in her early 30s, with a warm and composed demeanor that instantly puts others at ease $\cdots$ \item Shakoo Maku Baba: A man with a short beard and warm eyes, wearing a red sweater, blue jeans, and brown shoes $\cdots$ \end{itemize} \textbf{Scene(trigger+action):} \textit{Shakoo Maku Lulu} sits beside \textit{Shakoo Maku Mama} and \textit{Shakoo Maku Baba} smiling. \textbf{Note(style):} storybook style illustration, soft colors, for children aged 3-6. \end{tcolorbox}

To evaluate CharCom’s ability to maintain character consistency in multi-scene narratives, we construct a synthetic benchmark of 20 storylines generated using GPT-4o. Each narrative consists of five semantically coherent prompts \(\mathcal{P} = \{p_1, \dots, p_5\} \), centered on a unique protagonist represented by a single 512 $\times$ 512 reference portrait. All competing methods are evaluated under identical prompt settings. Although CharCom supports consistent generation for multiple characters, all primary quantitative evaluations and ablation studies are conducted on a single-character setting (\textit{Shakoo Maku Lulu}, the main child protagonist) unless explicitly stated otherwise. This choice facilitates fair comparison and simplifies metric computation, as single-character scenarios avoid ambiguity in identity matching.

We compare CharCom against three baselines: (i) \emph{Vanilla FLUX}, the original diffusion model without character conditioning, (ii) \emph{FLUX + IP-Adapter}, which incorporates image-based personalization, and (iii) \emph{Static-All}, a prompt-unaware LoRA baseline inspired by ORACLE~\cite{oracle2024oracle} that loads all character adapters with equal weights ($w_c=1$) at inference.  CharCom is evaluated in compositional inference mode without additional fine-tuning. This setting reveals potential interference that occurs when identity modules are injected without semantic disentanglement.

\begin{table}[htbp]
\centering
\small
\caption{Evaluation metrics for identity preservation, prompt alignment, and temporal coherence. $\mathcal{F}$: generated image, $\mathcal{R}$: reference portrait, $\mathcal{P}$: scene prompt.}
\label{tab:metrics_definition}
\begin{tabularx}{\linewidth}{>{\hsize=0.8\hsize}p{3.5cm}X}
\toprule
\textbf{Metric} & \textbf{Description} \\
\midrule
\textbf{\scriptsize Identity Score (IS)} &
Measures visual similarity between the generated image $\mathcal{F}$ and the reference portrait $\mathcal{R}$, scored in $[1,5]$. \\

\textbf{\scriptsize Prompt Fidelity Score (PFS)} &
Measures semantic alignment between the generated image $\mathcal{F}$ and the input prompt $\mathcal{P}$, scored in $[1,5]$. \\

\textbf{\scriptsize Integrated Consistency Score (ICS)} &
Measures overall consistency by combining IS and PFS via normalized product: $\text{ICS} = \frac{1}{25} \cdot \text{IS} \cdot \text{PFS}$, scaled to $[0.04,1]$. \\

\textbf{\scriptsize Temporal ICS (T-ICS)} &
Measures cross-frame identity consistency over $N$ frames using a VLLM-based temporal evaluator $\mathcal{T}$. \\

\textbf{\scriptsize T-ICS\textsubscript{Emb}} &
Measures facial-level consistency by averaging cosine similarity between adjacent face embeddings extracted from $\mathcal{F}$ by a lightweight face encoder. \\
\bottomrule
\end{tabularx}
\end{table}

All evaluations are conducted using GPT-4o as a VLLM under a triple-blind protocol. We employ five metrics to evaluate identity preservation, prompt alignment, and temporal coherence: Identity Score (IS), Prompt Fidelity Score (PFS), Integrated Consistency Score (ICS), Temporal ICS (T-ICS), and its face-aware variant, T-ICS\textsubscript{Emb}. Table~\ref{tab:metrics_definition} summarizes their descriptions and corresponding evaluation ranges. The formal definitions of these metrics are given as follows:
\[
\begin{gathered}
\text{IS} = \mathcal{V}_{\text{sim}}(\mathcal{R}, \mathcal{F}) \in [1,5]  \\
\text{PFS} = \mathcal{S}_{\text{align}}(\mathcal{P}, \mathcal{F}) \in [1,5] \\
\text{ICS} = \frac{1}{25} \cdot \text{IS} \cdot \text{PFS} \in [0.04,1] \\
\text{T-ICS} = \frac{1}{|\mathcal{C}|} \sum_{c \in \mathcal{C}} \left[ \frac{1}{N-1} \sum_{i=1}^{N-1} \mathcal{T}_c(\mathcal{F}_{i}, \mathcal{F}_{i+1}) \right] \\
\text{T-ICS}_{\text{Emb}} = \frac{1}{|\mathcal{C}|} \sum_{c \in \mathcal{C}} \left[ \frac{1}{N-1} \sum_{i=1}^{N-1} \cos\left( f_{\text{emb}}\big(\hat{\mathcal{F}}_{i}^{(c)}\big), f_{\text{emb}}\big(\hat{\mathcal{F}}_{i+1}^{(c)}\big) \right) \right]
\end{gathered}
\]
Here, \(\hat{\mathcal{F}}_i^{(c)}\) denotes the cropped face region of character $c$ in frame $i$, and \(f_{\text{emb}}(\cdot)\) denotes the embedding function of the face encoder.

\subsection{Experimental Results}

Most of the metrics are computed via GPT-4o as a VQA evaluator under a fixed scoring rubric. To ensure reliability, 30\% of the cases were randomly sampled for human review, showing strong agreement with GPT-4o's scores. To assess temporal consistency, we quantify how each method preserves character identity across sequential scenes using the metrics in Table~\ref{tab:joint_eval}. Vanilla FLUX, without any personalization, suffers the most severe drift. IP-Adapter mitigates drift via visual anchors, but it still struggles to maintain facial-level stability in dynamic narratives. Static-All improves cross-frame smoothness by loading all adapters simultaneously; however, it induces character confusion because unrelated modules inject conflicting features, e.g., attributes from the grandfather leaking into the father. This yields higher T-ICS but depressed IS/PFS, consistent with entangled identity control. In contrast, CharCom attains the highest temporal stability by selectively composing only relevant character adapters under structured prompts, thereby balancing narrative coherence with identity specificity and outperforming all baselines in anthropometric continuity (shown in Fig.~\ref{fig:teaser}).

\begin{table}[htbp]
\centering
\small
\caption{Quantitative comparison of identity consistency (IS/PFS/ICS) and temporal consistency (T-ICS/T-ICS\textsubscript{Emb}) across methods.}
\label{tab:joint_eval}
\begin{tabular}{lccc|cc}
\toprule
\textbf{Method} & \textbf{IS} $\uparrow$ & \textbf{PFS} $\uparrow$ & \textbf{ICS} $\uparrow$ 
& \textbf{T-ICS} $\uparrow$ & $\mathbf{T\text{-}ICS}_{\text{Emb}}$ \\
\midrule
\scriptsize{Flux (Vanilla)} & $2.84 \pm 0.16$ & $4.03$ & $0.31$ 
               & $0.27$ & 0.4432 \\
\scriptsize{Flux + IP-Adapter} & $3.91 \pm 0.13$ & $4.27$ & $0.55$ 
                 & $0.52$ & 0.5678 \\
\scriptsize{Static-All} & $3.41 \pm 0.22$ & $3.95$ & $0.58$ 
                 & $0.66$ & 0.5907 \\
\scriptsize{\textbf{CharCom (Ours)}} & $\mathbf{4.63 \pm 0.08}$ & $\mathbf{4.51}$ & $\mathbf{0.73}$ 
                           & $\mathbf{0.74}$ & $\mathbf{0.8742}$ \\
\bottomrule
\end{tabular}
\end{table}
Subjective human evaluations further corroborate the quantitative findings. As shown in Table~\ref{tab:human_eval}, CharCom attains the highest average rating (4.40), outperforming all baselines on identity, alignment, and narrative coherence. While IP-Adapter achieves decent prompt alignment, it underperforms on identity preservation. Static-All benefits from LoRA injection but suffers from perceptual confusion in coherence due to irrelevant adapters being activated. CharCom’s structured composition and prompt-aware control yield more coherent and believable narratives from a human perspective.

\begin{table}[htbp]
\centering
\small 
\caption{Human evaluation on character, alignment, and coherence}
\label{tab:human_eval}
\begin{tabular}{lccc|c}
\toprule
\textbf{Method} & {\footnotesize \textbf{Identity}} $\uparrow$ 
               & {\footnotesize \textbf{Alignment}} $\uparrow$ 
               & {\footnotesize \textbf{Coherence}} $\uparrow$ 
               & \textbf{Avg.} \\
\midrule
Flux (Vanilla) & 2.7 & 3.9 & 2.5 & 3.03 \\
Flux + IP-Adapter & 3.8 & 4.1 & 3.5 & 3.80 \\
Static-All & 3.6 & 4.0 & 2.2 & 3.27 \\
\textbf{CharCom (Ours)} & \textbf{4.6} & \textbf{4.4} & \textbf{4.2} & \textbf{4.40} \\
\bottomrule
\end{tabular}
\end{table}

We further examine each method’s scalability under increasing scene complexity (Table~\ref{tab:multi_character_scaling}). As the number of characters per scene grows from one to four, all methods degrade in both identity and temporal metrics. However, CharCom remains robust, exhibiting minimal identity drift and prompt misalignment under multi-character compositions. In contrast, IP-Adapter and Static-All decline more sharply, reflecting their limited capacity for disentangled control. Flux, lacking any personalization, fails to preserve distinct identities altogether. These trends highlight CharCom’s advantage in compositional generation, particularly in crowded or narrative-rich scenarios.

\begin{table}[htbp]
\centering
\small
\caption{Scalability to multi-character scenes. We gradually increase the number of involved characters from 1 to 4}
\label{tab:multi_character_scaling}
\begin{tabular}{lcccccc}
\toprule
\textbf{Method} & \textbf{\#Chars} & \textbf{IS} $\uparrow$ & \textbf{PFS} $\uparrow$ & \textbf{ICS} $\uparrow$ & \textbf{T-ICS} $\uparrow$ & $\mathbf{T\text{-}ICS}_{\text{Emb}}$ \\
\midrule
\multirow{4}{*}{\scriptsize Flux (Vanilla)} 
& 1 & 2.84 & 4.03 & 0.31 & 0.27 & 0.4432 \\
& 2 & 2.45 & 3.92 & 0.27 & 0.21 & 0.3041 \\
& 3 & 2.18 & 3.80 & 0.24 & 0.18 & 0.2694 \\
& 4 & 1.96 & 3.67 & 0.20 & 0.15 & 0.1311 \\
\midrule
\multirow{4}{*}{\scriptsize Flux + IP-Adapter} 
& 1 & 3.91 & 4.27 & 0.55 & 0.52 & 0.5678 \\
& 2 & 3.71 & 4.21 & 0.51 & 0.45 & 0.4301 \\
& 3 & 3.55 & 4.15 & 0.47 & 0.40 & 0.4141 \\
& 4 & 3.32 & 4.06 & 0.42 & 0.35 & 0.2898 \\
\midrule
\multirow{4}{*}{\scriptsize Static-All} 
& 1 & 3.88 & 4.18 & 0.51 & 0.51 & 0.5297 \\
& 2 & 3.62 & 4.19 & 0.50 & 0.39 & 0.4115 \\
& 3 & 3.34 & 4.07 & 0.46 & 0.31 & 0.3267 \\
& 4 & 2.98 & 4.01 & 0.36 & 0.22 & 0.2012 \\
\midrule
\multirow{4}{*}{\textbf{\scriptsize CharCom (Ours)}} 
& 1 & \textbf{4.63} & \textbf{4.51} & \textbf{0.73} & \textbf{0.74} & \textbf{0.8742} \\
& 2 & 4.52 & 4.47 & 0.71 & 0.70 & 0.7612 \\
& 3 & 4.37 & 4.41 & 0.69 & 0.67 & 0.7283 \\
& 4 & 4.21 & 4.33 & 0.66 & 0.64 & 0.6237 \\
\bottomrule
\end{tabular}
\end{table}

These findings underscore the robustness of CharCom’s compositional framework in preserving distinct character identities, even as narrative complexity increases.

\subsection{Ablation Study}

We perform ablation studies to assess the contribution of three key design components in CharCom: structured prompt design, adapter composition strategy, and the ordering of reference prompts.
\begin{table}[htbp]
\centering
\small
\caption{Ablation on prompt and adapter}
\label{tab:ablation_main}
\begin{tabular}{lccc}
\toprule
\textbf{Variant} & IS $\uparrow$ & T-ICS $\uparrow$ & $\mathbf{T\text{-}ICS}_{\text{Emb}}$ \\
\midrule
CharCom (full) & 4.63 & 0.74 & 0.8742 \\
w/o Structured Prompt & 3.71 & 0.58 & 0.5432 \\
w/o Adapter Composition & 3.96 & 0.69 & 0.6714 \\
Random Prompt Order & 4.11 & 0.63 & 0.6488 \\
\bottomrule
\end{tabular}
\end{table}

To further examine the contribution of each system component, we conducted targeted ablation experiments, as summarized in Table~\ref{tab:ablation_main}. Removing the structured prompt design caused a sharp performance drop in both IS (4.63 $\rightarrow$ 3.71) and T-ICS (0.74 $\rightarrow$ 0.58), highlighting the crucial role of well-structured prompt formulation in preserving character identity. Disabling the low-rank adapter composition mechanism also led to a clear decline in consistency scores, confirming the necessity of modular adapter injection for multi-character, multi-scene generation. Finally, randomizing the prompt order in multi-character settings further degraded performance, indicating that maintaining a semantically consistent order is essential for coherent spatial arrangement and effective character disentanglement.
\begin{table}[htbp]
\centering
\small
\caption{Impact of reference count on consistency}
\label{tab:ablation_fewshot}
\begin{tabular}{cccc}
\toprule
\textbf{Ref Count} & IS $\uparrow$ & T-ICS $\uparrow$ & $\mathbf{T\text{-}ICS}_{\text{Emb}}$ \\
\midrule
1  & 2.92 & 0.36 & 0.3119 \\
5  & 3.48 & 0.41 & 0.5760 \\
15 & 3.96 & 0.66 & 0.8104 \\
30 & 4.11 & 0.82 & 0.8742 \\
\bottomrule
\end{tabular}
\end{table}

We also examine how the quantity of reference images influences adapter training and, consequently, generation quality. We vary the number of portrait references per character across four settings (1, 5, 15, and 30), with results shown in Table~\ref{tab:ablation_fewshot}. As expected, more reference portraits consistently improve both identity and temporal consistency scores. Remarkably, even with only a single reference image, CharCom achieves competitive results, demonstrating strong adaptability in low-resource conditions and practical applicability to real-world creative workflows where character references may be scarce.

\section{Discussion}

We provide a comprehensive analysis of CharCom, covering its strengths, limitations, and representative failure cases, along with insights into potential future improvements.

\subsection{Strengths of the Framework}

\paragraph{Parameter-Efficient Scalability} 
Each LoRA adapter enables lightweight and scalable per-character adaptation. The base diffusion model remains frozen throughout training and inference. At inference time, CharCom supports plug-and-play adapter composition, enabling dynamic scene construction without retraining.
\paragraph{Compositional Flexibility} 
CharCom enables arbitrary character selection and dynamic adapter composition during inference, supporting modular and customizable multi-character generation. This design enables fine-grained control over character inclusion and prominence, without incurring additional computational overhead.
\paragraph{Robust Prompt Generalization} 
Experimental results show that CharCom consistently preserves character fidelity across diverse prompt variations, including changes in style, emotion, and scene structure. This robustness arises from CharCom’s integration into the generation pipeline, which stabilizes character embeddings under both visual and linguistic variations.

Collectively, these strengths make CharCom a practical, scalable, and parameter-efficient solution for real-world multi-character generation.

\subsection{Limitations and Failure Modes}

Despite its strong performance, CharCom exhibits several limitations and failure cases in specific scenarios.
\begin{figure}[htbp]
  \centering
  \begin{subfigure}[b]{0.48\columnwidth}
    \centering
    \includegraphics[width=0.75\linewidth]{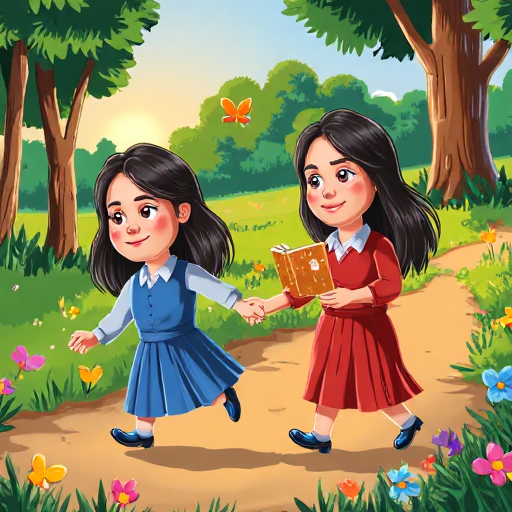}
    \caption{Identity disambiguation.}
    \label{fig:mistake1a}
  \end{subfigure}
  \hfill
  \begin{subfigure}[b]{0.48\columnwidth}
    \centering
    \includegraphics[width=0.75\linewidth]{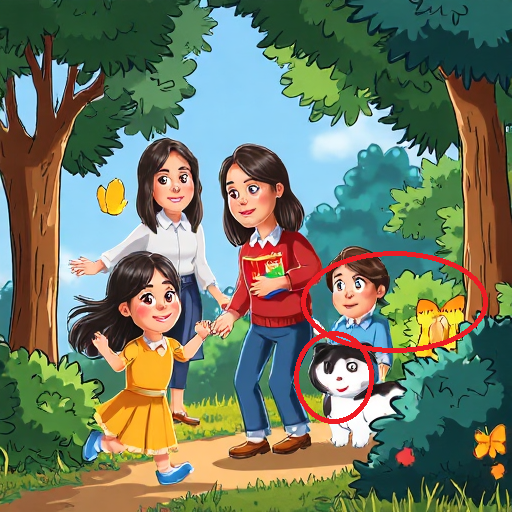}
    \caption{Attention interference.}
    \label{fig:mistake1b}
  \end{subfigure}
  \caption{Two representative failure cases: Ambiguous prompts (e.g., “two girls”) lead to character confusion. (b) Peripheral characters lose fidelity in crowded scenes due to attention interference.}
  \Description{Two side-by-side images: left shows character confusion, right shows degradation in peripheral characters.}
  \label{fig:combined_failures}
\end{figure}

\paragraph{Ambiguity-Induced Identity Drift} 
As shown in Fig.~\ref{fig:mistake1a}, prompts involving multiple visually similar characters can trigger attention blending, causing merged or ambiguous identities. Such cases frequently occur in symmetric or crowded scenes lacking explicit spatial or role disambiguation.
\paragraph{Peripheral Fidelity Degradation from Attention Bias} 
Fig.~\ref{fig:mistake1b} illustrates a hierarchical attention bias: central characters attract disproportionate attention, while peripheral figures suffer quality degradation. This suggests an attention capacity bottleneck when multiple LoRA modules are active concurrently.

These observations point to promising directions for future work, including improved spatial reasoning, scalable attention mechanisms, and more robust prompt disambiguation—critical for handling increasingly complex and densely populated scenes.

\section{Conclusion}

We present \textbf{CharCom}, a lightweight and modular framework for character-consistent story illustration. By representing each character with a compact LoRA adapter, CharCom enables plug-and-play composition without retraining, supporting scalable multi-character generation. Experimental results on a multi-scene benchmark demonstrate significant gains over prior methods in identity preservation, prompt alignment, and temporal coherence. While challenges remain in disambiguating similar characters and balancing attention in crowded layouts, CharCom offers a practical and extensible foundation for story-level generation.

\bibliographystyle{ACM-Reference-Format}
\bibliography{sample-base}

\end{document}